\pdfoutput=1

\documentclass[11pt]{article}
\usepackage{misc/acl}
\usepackage{times}
\usepackage{latexsym}
\usepackage[T1]{fontenc}
\usepackage[utf8]{inputenc}
\usepackage{microtype}
\usepackage{todonotes}
\usepackage{amsmath}
\usepackage{amssymb}
\usepackage{cleveref}
\usepackage{booktabs}
\usepackage{lipsum}
\usepackage{enumitem}
\usepackage{xspace}
\usepackage{listings}
\usepackage{soul} %
\usepackage{caption}
\usepackage{transparent} %
\usepackage{tcolorbox} %
\usepackage{multirow}
\usepackage{ulem}
\usepackage{wasysym} %

\crefname{lstlisting}{listing}{listings}
\Crefname{lstlisting}{Listing}{Listings}
\crefname{equ}{equation}{equations}
\Crefname{equ}{Equation}{Equations}

\DeclareCaptionType{equ}[Equation][List of equations]

\usepackage{courier}
\usepackage{marginnote}

\definecolor{TodoColor}{rgb}{1,0.7,0.6}

\sethlcolor{TodoColor}

\definecolor{DocumentLinkColor}{rgb}{0.4,0.6,0.3}
\newcommand{\hrefEmail}[2]{\href{mailto:#1}{\color{black}{#2}}}

\newcommand{\researchquestionbox}[1]{
    {
    \vspace{-1.2mm}
    \begin{tcolorbox}[colback=white!0,boxrule=0.9pt,right=2mm,left=2mm,bottom=2mm,top=2mm]
        \fontsize{10pt}{10pt}\selectfont
        \vspace{-1mm} #1
        
        \vspace{-0.8mm}
    \end{tcolorbox}
    }
    \vspace{-2mm}
}

\newcommand{\bpeatsymbol}{\hspace{-0.4mm}@\hspace{-1mm}@\hspace{1mm}}

\title{Stolen Subwords: Importance of Vocabularies for\\Machine Translation Model Stealing}

\author{Vilém Zouhar\\
  \texttt{\hrefEmail{vzouhar@ethz.ch}{vzouhar@ethz.ch}}
}

\begin{document}
\maketitle

\begin{abstract}
In learning-based functionality stealing, the attacker is trying to build a local model based on the victim's outputs.
The attacker has to make choices regarding the local model's architecture, optimization method and, specifically for NLP models, subword vocabulary, such as BPE.
On the machine translation task, we explore (1) whether the choice of the vocabulary plays a role in model stealing scenarios and (2) if it is possible to extract the victim's vocabulary.
We find that the vocabulary itself does not have a large effect on the local model's performance.
Given gray-box model access, it is possible to collect the victim's vocabulary by collecting the outputs (detokenized subwords on the output).
The results of the minimum effect of vocabulary choice are important more broadly for black-box knowledge distillation.
\end{abstract}

\footnotetext[0]{\href{https://github.com/zouharvi/vocab-stealing}{github.com/zouharvi/vocab-stealing}}

\section{Introduction}

NLP models are a key intellectual property, many of which are deployed online.
This access creates an attack surface by which an adversarial agent can attempt to replicate the model at the fraction of the cost of the original model training or with the absence of proper training data.
Their task, called model functionality stealing, is to create a local copy (student) of a model (victim/teacher).
In learning-based approaches, the victim model is queried to create a synthetic dataset, on which the student model is trained.
Specifically for machine translation from $L_1$ to $L_2$, the attacker has access to a monolingual dataset $\mathcal{D}^{L_1}$, creates a synthetic dataset in $L_2$ by querining the victim and trains a model on $(\mathcal{D}^{L_1}, M_V(\mathcal{D}^{L_1}))$.
This is similar to model distillation \citep{Hinton2015DistillingTK,Freitag2017EnsembleDF,Wei2019OnlineDF,Tan2019MultilingualNM,Gordon2020DistillAD,zhou2021distilling}, though it usually includes full model access and the goal of making the student model more efficient or privacy-preserving.

In the model stealing scenario, the attacker has to make decisions regarding the student model: data used for training $D^{L_1}$, model architecture and other preprocessing, including the subword algorithm and its vocabulary.
While the effect of data and architecture choice in distillation and model stealing distillation has been explored \citep{orekondy2019knockoff,krishna2019thieves,wallace2020imitation,zouhar2021sampling}, it is unclear what choice should be made regarding the model vocabulary.
Specifically, we focus on BPE \citep{shibata1999byte,Sennrich2016NeuralMT}, a popular subwording algorithm \citep{ding2019call}.
We wish to quantify how much the increased student performance (gained by having access to the victim vocabulary) is worth the cost of obtaining such vocabulary.
This leads to questions on vocabulary importance (\textbf{RQ1}) and vocabulary inference (\textbf{RQ2}):

\researchquestionbox{
\textbf{RQ1}: Is it advantageous for learning-based model stealing to know the victim's BPE vocabulary or is domain-specificity more important? \\
\textbf{RQ2}: Is it possible to recover the victim's vocabulary given black-box and gray-box access?
}

We describe the model stealing setup in \Cref{sec:model_stealing} and the BPE algorithm in \Cref{sec:bpe}.
In \Cref{sec:model_stealing_results} we examine the task of learning-based model stealing and compare the effect of vocabularies on the student's performance (\textbf{RQ1}).
Here we find that training on the victim's vocabulary is marginally worse than using a BPE vocabulary trained on the relevant domain, better than unrelated domains and when trained all data.
In \Cref{sec:recovering_victims_bpe_vocabulary} we describe approaches for recovering the victim's BPE vocabulary based on the level of access and local data (\textbf{RQ2}).
We discuss the results and conclude in \Cref{sec:discussion,sec:conclusion}.
Note the discussion on Limitations and Ethics in \Cref{sec:limitations,sec:ethics}.
For clarity, we show all results in the main paper as averages of BLEU on English $\leftrightarrow$ German language direction but plan to replicate the main findings for other languages and with other evaluation metrics.

\section{Related Work}

Model stealing has been explored mostly in the domain of computer vision where the output is a classification, possibly with probabilities \citep{tramer2016stealing,orekondy2019knockoff,atli2020extraction,kariyappa2021maze,szyller2021good,liu2022stolenencoder}.
However, recent works have shown that even in NLP, where the output domain is more complicated, it is possible to infer training data \citep{carlini2021extracting} and some weights \citep{zanella2021grey} of large language models.
It is also possible to do learning-based model stealing of such models \citep{krishna2019thieves,keskar2020thieves,lyu2021killing,xu2022student}.

Most work in this area for MT has been framed as knowledge distillation.
If KL-divergence is to be used, it needs to be computed over matching distributions which assumes the same subword vocabulary.
For MT distillation and immitation attacks, even without token-level KL-divergence optimization, \citet{Freitag2017EnsembleDF,Wei2019OnlineDF,Tan2019MultilingualNM,Gordon2020DistillAD} implicitly use the same BPE vocabulary as the teacher while \citep{wallace2020imitation,zouhar2021sampling} choose to train their own BPE without any justification.
The distillation work also dealt with the problem of mismatched student and teacher vocabularies \citep{khanuja2021mergedistill,kolesnikova2022knowledge}.

\citet{ding2019call,gowda2020finding} examine the effect of BPE vocabulary size and \citet{bogoychev2021highs} experiment with using BPE trained on a different domain and is therefore suboptimal for the primary one.
Tokenization of the training data is well-known to affect machine translation and other NLP model performance \citep{howmuchtokenization,howmuchtokenization2,zouhar-etal-2023-tokenization}.

\section{Model Stealing}
\label{sec:model_stealing}

For most experiments, we consider black-box access (only translated outputs are available) to the victim model $M_V$.
To answer \textbf{RQ2} we use additional grey-box access where the output is still segmented to subwords.\footnote{Available e.g. by GPT-3 \citep{brown2020language}.}
See \Cref{tab:blackbox_graybox_example} for an example of black-box and gray-box outputs.

While important in most adversarial scenarios, we assume a large but fixed query budget for the model of 10M sentence queries, in order to translate the respective datasets.
For the victim model, we use the winning WMT19 English$\leftrightarrow$German model \citep{ng2019facebook} as $M_V$.
We translate all the available authentic data $\mathcal{D}_{A}$ using the victim model: $\mathcal{D}_{V}^{L_1} = M_V(\mathcal{D}_{A}^{L_2})$.
We then train several student models on a combination of this data with various BPE models.

\paragraph{Student model description.}

The student model is Transformer-based on the Fairseq configuration \texttt{transformer\_iwslt\_de\_en}.
Following the victim model \citep{ng2019facebook}, we use a joint BPE vocabulary of 30k entries for the student model in order to reduce the number of configurations.

\paragraph{Data.}

We use 10M parallel sentences from the following datasets: ParaCrawl \citep{espla2019paracrawl,banon2020paracrawl}, EuroPat \citep{heafield2022europat} and CommonCrawlAligned \citep{elkishky_ccaligned_2020} to which we refer to as PCrawl, EuroPat and CCrawl, respectively.
They were chosen to represent different domains: general, legal publications and patents.
For each of these datasets, we use 9.8M, 100k and 100k for the training, development and test splits.
See \Cref{tab:data_example,tab:data_overview} for dataset overview and example sentences and translations.

\section{Byte-Pair-Encoding}
\label{sec:bpe}

BPE is a method to reduce the output dimensionality by splitting target words into smaller units (subwords).
This way, the MT system does not have to model the word \emph{moonlight}, which occurs in our data 272 times, independently but rather model \emph{moon\bpeatsymbol} and \emph{light} which occur 7076 and 587321 times, respectively.
See \Cref{tab:blackbox_graybox_example} for an example of subword units.

\paragraph{Vocabulary efficiency.}

BPE is a compression algorithm and as such, its goal is to encode the data in as little space as possible (measured by expected token count).
Because BPE encodes the data as a sequence of subwords, we consider the number of subwords needed to encode a given dataset.
With the same fixed vocabulary size (30k), we train multiple BPE models $B_i$ on datasets $\mathcal{D}_i$ and define the efficiency of this model on dataset $\mathcal{D}_j$ as $\frac{|B_i(\mathcal{D}_j)|}{|B_j(\mathcal{D}_j)|}$.
This is the number of subwords needed to encode $\mathcal{D}_j$ with BPE model $B_i$ divided by the space requirements of the most efficient model with the same vocabulary size constraints ($B_j$).

We show the BPE model efficiencies on the three datasets in \Cref{tab:bpe_vocab_efficiency}.
The efficiencies confirm that BPE models trained on different datasets, even on the same language, are suboptimal on different domains.
It also shows that PCrawl and CCrawl contain similar distribution (scrapped from the web).

\begin{table}[htbp]
\centering
\resizebox{0.85\linewidth}{!}{
\begin{tabular}{clccc}
\toprule
& & \multicolumn{3}{c}{\textbf{Target dataset}} \\
& & PCrawl & CCrawl & EuroPat \\ 
\cmidrule{2-5}
\parbox[t]{2mm}{\multirow{5}{*}{\rotatebox[origin=c]{90}{\textbf{Training}}}}
& Victim & 1.08 \hspace{1.8mm} & 1.10 \hspace{1.8mm} & 1.30 \hspace{1.8mm} \\ 
& All & 1.04 \hspace{1.8mm} & 1.04 \hspace{1.8mm} & 1.05 \hspace{1.8mm} \\ 
& PCrawl & 1.00 $\star$ & 1.03 \hspace{1.8mm} & 1.26 \hspace{1.8mm} \\ 
& CCrawl & 1.04 \hspace{1.8mm} & 1.00 $\star$ & 1.28 \hspace{1.8mm} \\ 
& Patents & 1.29 \hspace{1.8mm} & 1.31 \hspace{1.8mm} & 1.00 $\star$ \\ 
\bottomrule
\end{tabular}
}
\caption{Efficiency of BPE encoding on various datasets in terms of the required number of subwords (lower is better). The ratio is computed against optimal BPE on that dataset (minimum in column, marked with star). \emph{Victim} was trained on a larger version of PCrawl.}
\label{tab:bpe_vocab_efficiency}
\end{table}

\section{Model Stealing Results}
\label{sec:model_stealing_results}

In learning-based model stealing, we first translate the available data by the victim's model and then train a student model on this data.
In \Cref{tab:model_aa_av_va_vv} we show that training on authentic parallel data (Auth.+Auth.) is the most optimal but training on the victim's outputs (Auth.+Victim) is not far behind ($-0.07$).

\begin{table}[htbp]
\centering
\resizebox{\linewidth}{!}{
\begin{tabular}{lcccc}
\toprule
\bf Source+Target & \bf PCrawl & \bf CCrawl & \bf EuroPat & \bf Avg.\\ 
\midrule
Auth.+Auth. & 35.07 & 36.42 & 35.13 & 35.54 \\

Auth.+Victim & 34.95 & 36.25 & 35.20 & 35.47 \\

\bottomrule
\end{tabular}
}
\caption{Within domain BLEU performance of student models trained on either authentic or victim's data (source and/or target). The used BPE is trained on the same data as the student model.}
\label{tab:model_aa_av_va_vv}
\end{table}

\begin{table}[htbp]
\centering
\resizebox{\linewidth}{!}{
\begin{tabular}{lcccc}
\toprule
\bf BPE & \bf PCrawl & \bf CCrawl & \bf EuroPat & \bf Avg. \\ 

\midrule
Victim & 34.73 & 36.38 & 34.16 & 35.09 \\
All & 34.91 & 36.05 & 34.36 & 35.11 \\
PCrawl & 34.95 & 35.74 & 34.23 & 34.97 \\
CCrawl & 34.86 & 36.25 & 34.19 & 35.10 \\
EuroPat & 34.54 & 35.82 & 34.41 & 34.92 \\
\bottomrule
\end{tabular}
}
\caption{Within domain student performance (BLEU) trained on Authentic + Victim data. The student models are trained with different vocabularies (first column).}
\label{tab:bpe_a_s_v}
\end{table}

In \Cref{tab:bpe_a_s_v} we show that training with the victim's vocabulary is marginally worse than training on the target domain ($-0.11$), beter on unrelated domains ($+0.23$) and similar when trained on all data ($+0.01$).
The BPE optimality provides some explanation for student performance trained with the specific vocabulary.
The higher the BPE inefficiency (further from optimal 1.00, \Cref{tab:bpe_vocab_efficiency}), the lower the student BLEU.
This is verified with negative correlations (Pearson: -58.27\%, p = 0.023, Spearman: -57.20\%, p = 0.026).\footnote{Computed as correlation between mean-normalized BLEU scores for each domain and its BPE efficiency. Using scipy 1.9.3 \citep{virtanen2020scipy}.}
Therefore, for model stealing and knowledge distillation applications it, therefore, plays only a trivial role to match the vocabulary of the local student model to that of the victim's (\textbf{RQ1}).

\begin{figure}[htbp]
\centering
\includegraphics[width=\linewidth]{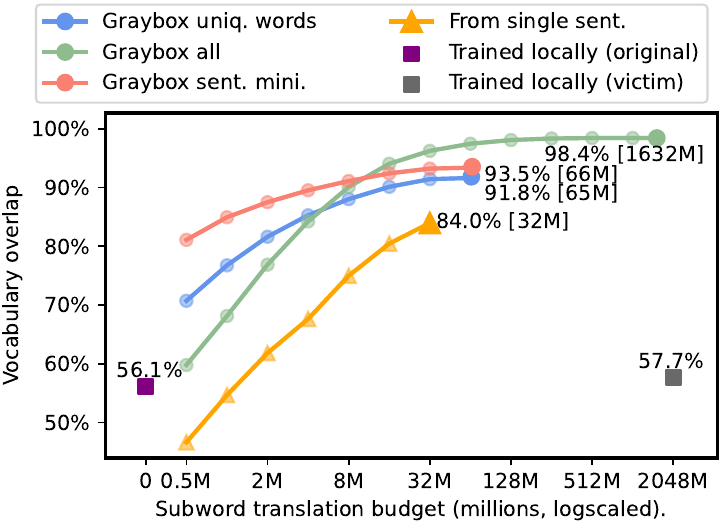}
\vspace{-5mm}
\caption{Overlap with victim's vocabulary. Numbers in square brackets indicate the subword translation budget. Points $\blacksquare$ are a BPEs trained locally, $\CIRCLE$ are vocabularies collected from model output and $\blacktriangle$ are vocabularies collected from model output starting from a single sentence (max of 5).}
\label{fig:vocab_budget}
\end{figure}

\section{Recovering Victim's BPE Vocabulary}
\label{sec:recovering_victims_bpe_vocabulary}

We operationalize the efficiency of recovering the victim's BPE vocabulary $V$ against a hypothesis vocabulary $V'$ as overlap $\frac{2\cdot |V \cap V'|}{|V|+|V'|}$ which ranges from 0 to 100\%.
We consider multiple baseline approaches and show their performance in \Cref{fig:vocab_budget}.
The \emph{Overlap} is not the only important variable but also \emph{Subword budget} (how many subwords does the victim need to translate) and whether we assume black-box or gray-box access.

\paragraph{Training locally.}
First, we can train our own BPE model on the local authentic data.
This does not require querying of the victim model but is sensitive to the original data selection and yields only $56.1\%$ overlap with the victim's vocabulary.
We can also train a BPE model on the victims' output.
This requires the victim to translate the whole data, making it the most expensive.
It is only marginally better ($57.7\%$) than training on authentic data.
This can be explained by the fact that the victim's BPE was trained on authentic and not synthetic data.
The difference in data distributions between the authentic and the translated data is shown by the number of unique tokens (13.7M and 4.5M, respectively).

\paragraph{Gray-box access.}
Assuming that the model produces subword outputs (see \Cref{tab:blackbox_graybox_example}), we can simply collect all the subwords that appear in the victim output.
BPE is constructed by merging the most co-occurring pairs which leads to the selected subwords that are added to the vocabulary to be frequent in the training corpora.
The overlap of $98.4\%$ with the victim's vocabulary is therefore not surprising.
However, this approach is severely inefficient.
Consider the two English sentences and their translations into German in \Cref{tab:uniq_vocab_example}.
We are spending the budget on repeatedly translating the same words which yield the same subwords on the output.
Because we only care about the set of subwords in the output, we can simply translate all unique words in our dataset individually with a much smaller budget (65M).
This results in a vocabulary that has $92.1\%$ overlap.
The reason for this degradation is that words are translated differently and hence into different subwords.\footnote{E.g. \emph{I have} $\rightarrow$ \emph{Ich habe} and \emph{They have} $\rightarrow$ \emph{Sie haben}.}
A compromise between translating whole sentences and unique words is to translate sentences but remove already-seen words.
This shows promising results, especially within the low-budget area.
A more elaborate approach with negative results is described in \Cref{sec:unique_small_min}.
All the observation-based gray-box approaches start to plateau from some point on because those subwords become rarer.
In \Cref{sec:missing_subwords} we analyze which subwords are missing.

\begin{table}[htbp]
\centering
\resizebox{0.9\linewidth}{!}{
\begin{tabular}{p{8cm}}
\toprule
The washing \uline{machine} is \uline{broke}n. \\
Die Wasch\bpeatsymbol \uline{maschine} ist \uline{kap\bpeatsymbol} utt\bpeatsymbol . \\
\midrule
I \uline{broke} the milling \uline{machine}. \\
Ich habe die F\bpeatsymbol rä\bpeatsymbol s\bpeatsymbol \uline{maschine} \uline{kap\bpeatsymbol} utt ge\bpeatsymbol macht\bpeatsymbol . \\
\bottomrule
\end{tabular}
}
\caption{Example gray-box translation with overlap.}
\label{tab:uniq_vocab_example}
\end{table}

\paragraph{From single sentence.}

Finally, we consider a scenario in which the attacker has not even monolingual data apart from a handful of sentences.
We use these sentences as starting seeds to a vocabulary and then create ``nonsense'' sequences of words sampled from this vocabulary.
We then translate this sequence, add resulting words into the opposing language's vocabulary, sample a ``nonsense'' sequence from that vocabulary and repeat.
To encourage sampling of less-sampled words, we sample a word with the weight of $\frac{1}{\text{\# sampled}}$.
We repeat this process until no changes to either the source or target vocabularies are made (patience of 5 iterations).
The algorithm is formally described in \Cref{lst:cyclic_vocab}.
While the results in \Cref{fig:vocab_budget} show that this method performs worse in recovering the victim's vocabulary, it should be appreciated in the context of starting with a single sentence.
The 5 starting sentences are shown in \Cref{tab:from_single_sent_sentences} and \Cref{fig:cycle_budget} shows that regardles of the starting seed, all runs converge to a similar vocabulary.

\paragraph{Take-away.}
The difficulty of inferring the victim's BPE vocabulary greatly depends on the level of access
If only black-box access is available, then training a BPE on the available authentic data is the best option.
If, however, the victim model produces subword outputs, simply collecting the output subwords makes it possible to construct a local copy of the vocabulary.
Contrary to intuition, it is more effective to query the model on authentic (possibly minified) sentences and not on single deduplicated words.
In the case where no authentic data is available, it is still possible to ineffectively infer a part of the vocabulary from a single seed sentence and ``nonsense'' sequence translations.

\section{Discussion}
\label{sec:discussion}

We show that the choice of the BPE vocabulary is largely inconsequential as long as (1) it is that of the victim or (2) it matches the student's domain.
Therefore the choice to either use the same vocabulary \citep{Freitag2017EnsembleDF,Wei2019OnlineDF,Tan2019MultilingualNM,Gordon2020DistillAD} or train a local one \citep{wallace2020imitation,zouhar2021sampling} is justified.

\section{Conclusion}
\label{sec:conclusion}

In this paper, we explored the setting of learning-based model stealing and focused on the issue of BPE subword vocabularies.
We find that it plays a very minor role whether the student shares the same vocabulary as the victim.
We also document several approaches for inferring the victim's vocabulary based on its outputs.
For both low and high translation budgets (0.5M and 128M) it is possible to infer the victim's vocabulary with the overlap of $93.5\%$ and $98.1\%$, respectively.

\vspace{-2mm}
\section*{Limitations}
\label{sec:limitations}
\vspace{-2mm}

We explored a specific NLP scenario (machine translation) and subwording method (BPE).
While we believe that the exploration of other subwording methods, such as Byte-level BPE \citep{wang2020neural}, would yield similar results, more tokenization-sensitive tasks could show higher dependency on victims's vocabulary.

\section*{Ethics statement}
\label{sec:ethics}

In our experiments we worked with publicly released models which we treated as belonging to the victim.
Our research does not advocate for unlawful stealing of intellectual propety, does not facilitate it, or provide any guidance.
The exception is the vocabulary extraction, which is not a key intellectual property component.

\bibliography{misc/bibliography.bib}

\clearpage
\appendix
\section{Error Analysis of Missing Subwords}
\label{sec:missing_subwords}

For the gray-box collection of subwords in translated sentences with 98.4\% overlap, there were 139 subwords missing.
The longest were obivous artefacts of the used dataset (such as \emph{www.nachrichten.at}).
However, for some subwords, such as \emph{Bundesligist}, the model had similar subwords: \emph{Bundeslig\bpeatsymbol} (which would be combined with \emph{ist}), \emph{Bundesliga} and \emph{Bundesligisten}.
Despite the word \emph{Shakespeare} appearing in the query data, on the output it was always presented as a single word (because of its frequency) and therefore the subword \emph{Shakes} was not found.
This also provides an explanation for the failure of the approach in \Cref{sec:unique_small_min}.

\section{Unique Word Minimization}
\label{sec:unique_small_min}

Despite taking only unique words, we are still duplicating work by translating both e.g. by translating both \emph{understand} $\rightarrow$ \emph{verstehen} and \emph{misunderstand} $\rightarrow$ \emph{miss\bpeatsymbol verstehen}.
We can therefore further minimize the budget by not querying words which are subwords of ones that we query later.
This is not an optimal approach because the composition of the larger word may be frequent enough to warrant its own subword, but still, this approach reaches $91.8\%$ overlap with 53M subword budget.
It is only marginally better than only taking unique words ($91.8\%$ with 65M, \Cref{fig:vocab_budget}).

\begin{lstlisting}[caption=Cyclic backtranslation to steal victim's vocabulary from a single sentence.,label=lst:cyclic_vocab,language=Python,escapeinside=`']
def translate_random(model, vocab):
    sent_src `$\leftarrow$' sample(vocab, k=20)
    sent_tgt `$\leftarrow$' model(sent_src)
    return {w | w `$\in$' sent_tgt}

SENT_0 `$\leftarrow$' "Alice was beginning to..."

vocab_en `$\leftarrow$' {w | w `$\in$' SENT_0}
vocab_de `$\leftarrow$' {}

`\textbf{loop}':
    vocab_de_ext `$\leftarrow$' translate_random(
        model_ende, vocab_en
    )
    if vocab_de_ext `$\subseteq$' vocab_de:
        `\textbf{exit}'
    vocab_de `$\leftarrow$' vocab_de `$\cup$' vocab_de_ext
    
    vocab_en_ext `$\leftarrow$' translate_random(
        model_deen, vocab_de
    )
    if vocab_en_ext `$\subseteq$' vocab_en:
        `\textbf{exit}'
    vocab_en `$\leftarrow$' vocab_en `$\cup$' vocab_en_ext
\end{lstlisting}

\begin{table}[htbp]
\centering
\resizebox{\linewidth}{!}{
\begin{tabular}{lccc}
\toprule
\textbf{Property} & \textbf{PCrawl} & \textbf{EuroPat} & \textbf{CCrawl} \\
\midrule
Line length (tokens) & 17 & 30 & 11 \\
Line length (chars) & 97 & 183 & 61 \\
Unique tokens & 4.6M & 4.1M & 3.2M \\
\hfill English & 2.2M & 0.9M & 1.7M \\
\hfill German & 3.7M & 3.5M & 2.7M \\
Unique chars & 6721 & 520 & 6897 \\
\hfill English & 6097 & 335 & 6062 \\
\hfill German & 4817 & 456 & 6335 \\
\bottomrule
\end{tabular}
}
\caption{Overview of words and characters found in the three used datasets (10M sentences each). Per-language line length distribution is not shown as it is similar. Tokens are compared case-insensitive. The datasets differ not only in the diversity of used words but also symbols (characters).}
\label{tab:data_overview}
\end{table}

\begin{table}[htbp]
\centering
\resizebox{\linewidth}{!}{
\begin{tabular}{lp{7cm}}
\toprule
\textbf{Input:} &
    Stolen Subwords: Importance of Vocabularies for Machine Translation Model Stealing \\
\midrule
\textbf{Black-box:} \hspace*{-3mm} &
    Gestohlene Subwörter: Bedeutung von Vokabeln für den Diebstahl maschineller Übersetzungsmodelle \\
\textbf{Gray-box:} \hspace*{-3mm} &
    Gest\bpeatsymbol ohl\bpeatsymbol ene Sub\bpeatsymbol wör\bpeatsymbol ter\bpeatsymbol : Bedeutung von V\bpeatsymbol ok\bpeatsymbol ab\bpeatsymbol eln für den Diebstahl masch\bpeatsymbol in\bpeatsymbol eller Über\bpeatsymbol setz\bpeatsymbol ungs\bpeatsymbol modelle \\
\bottomrule
\end{tabular}
}
\caption{Examples of black-box vs gray-box outputs from English to German translation.}
\label{tab:blackbox_graybox_example}
\end{table}

\begin{table}[htbp]
\centering
\resizebox{\linewidth}{!}{
\begin{tabular}{p{8cm}}
\toprule
- Stolen subwords: importance of vocabularies for machine translation model stealing \\
- NLP models are a key intelectual property, many of which are deployed online. \\
- We present the Eyetracked Multi-Modal Translation (EMMT) corpus, a dataset containing monocular eye movement recordings, audio and 4-electrode electroencephalogram (EEG) data of 43 participants. \\
- Two roads diverged in a wood, and I- I took the one less traveled by, And that has made all the difference. \\
- One morning, when Gregor Samsa woke from troubled dreams, he found himself transformed in his bed into a 
horrible vermin. \\ 
\bottomrule
\end{tabular}
}
\caption{Starting sentences to seed the vocabulary in \Cref{lst:cyclic_vocab}.}
\label{tab:from_single_sent_sentences}
\end{table}

\begin{figure}[htbp]
\centering
\includegraphics[width=\linewidth]{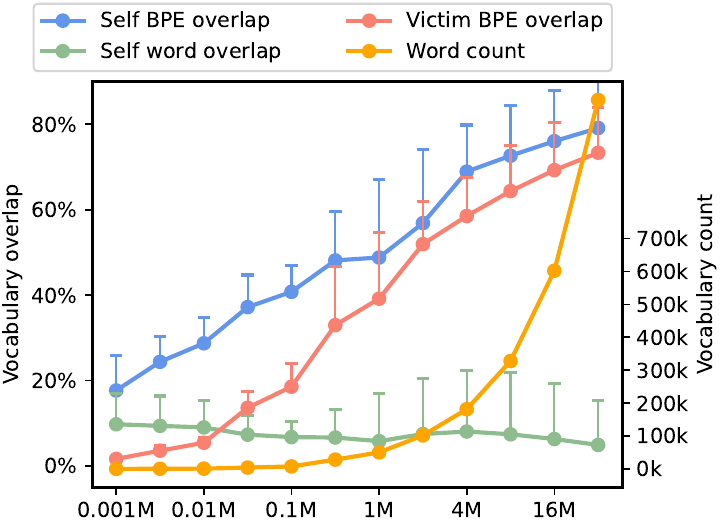}
\caption{Overlaps between vocabularies of different runs of the cyclic translations from single sentence algorithm (\Cref{lst:cyclic_vocab}) and vocabulary sizes. Points show the average of 5 seeds and bars the maximum.}
\label{fig:cycle_budget}
\end{figure}

\begin{table*}[htbp]
\centering
\resizebox{\linewidth}{!}{
\begin{tabular}{p{1.5cm}cp{7cm}p{7.6cm}}
\toprule
\textbf{Dataset} & \textbf{Source} & \textbf{English} & \textbf{German} \\
\midrule
PCrawl ParaCrawl &
    Original &
    Airbnb® | Devala - Holiday Rentals \& Places to Stay - Tamil Nadu, India &
    Airbnb® | Kuchanur – Ferienwohnungen \& Unterkünfte - Tamil Nadu, Indien \\
&
    Translated &
    Airbnb | Kuchanur Apartments \& Accommodations - Tamil Nadu, India &
    Airbnb ® | Devala - Ferienwohnungen und Unterkünfte - Tamil Nadu, Indien \\
\cmidrule{1-4}
EuroPat &
    Original &
    With this positioning, residual water can flow out of the chambers of the differential-pressure fluid gauge chamber due to the effect of gravity &
    Bei dieser Positionierung kann Restwasser aus den Kammern der Differenzdruckdose unter Schwerkraftwirkung abfließen. \\
&
    Translated &
    During this positioning, residual water can drain out of the chambers of the differential pressure box under gravity. &
    Bei dieser Positionierung kann aufgrund der Schwerkraft Restwasser aus den Kammern der Differenzdruck-Fluidmanometer-Kammer fließen. \\
\cmidrule{1-4}
CCrawl CCAligned &
    Original &
    Those who do good, good will be done to them! &
    Wer Gutes tut, dem wird Gutes widerfahren! \\
&
    Translated &
    He who does good will do good! &
    Diejenigen, die Gutes tun, werden Gutes tun! \\
\bottomrule
\end{tabular}
}
\caption{Example sentences from the three used English-German datasets with different domains. Rows marked with \emph{Translated} contain the victim's output.}
\label{tab:data_example}
\end{table*}

\end{document}